\documentclass[runningheads]{llncs}
\usepackage[T1]{fontenc}
\usepackage{graphicx}
\usepackage{svg}
\usepackage{epsfig}
\usepackage[numbers]{natbib}
\usepackage{doi}
\usepackage{amsmath}
\usepackage{mathtools}

\usepackage{microtype}
\usepackage{gensymb}
\usepackage{wrapfig}
\usepackage{amsfonts}       %
\usepackage{multicol}
\usepackage{multirow}
\usepackage{tabulary}
\usepackage{ctable}

\usepackage[utf8]{inputenc} %
\usepackage{booktabs}       %
\usepackage{nicefrac}       %

\usepackage{lipsum}		%
\usepackage{array}
\newcolumntype{L}[1]{>{\raggedright\let\newline\\\arraybackslash\hspace{0pt}}m{#1}}
\newcolumntype{C}[1]{>{\centering\let\newline\\\arraybackslash\hspace{0pt}}m{#1}}
\newcolumntype{R}[1]{>{\raggedleft\let\newline\\\arraybackslash\hspace{0pt}}m{#1}}
\usepackage{tabulary}
\usepackage{ctable}
\usepackage{subcaption}
\usepackage{amssymb}

\begin{document}
\title{CLIP-VQDiffusion : Langauge Free Training of Text To Image generation using CLIP and vector quantized diffusion model}
\titlerunning{CLIP-VQDiffusion}
\author{Seung Dae Han\inst{1}\orcidID{0009-0006-3897-6270} \and
Joo hee Kim\inst{1}\orcidID{0009-0003-8190-0827}}
\authorrunning{S. Han, J. Kim.}
\institute{INFINIQ, 459-11, Gasan-dong, Geumcheon-gu, Seoul, KOREA\\
\email{\{sdhan,jhkim\}@infiniq.co.kr}}  
\maketitle              %
\begin{abstract}
There has been a significant progress in text conditional image generation models. Recent advancements in this field depend not only on improvements in model structures, but also vast quantities of text-image paired datasets. However, creating these kinds of datasets is very costly and requires a substantial amount of labor. Famous face datasets don't have corresponding text captions, making it difficult to develop text conditional image generation models on these datasets. Some research has focused on developing text to image generation models using only images without text captions. Here, we propose CLIP-VQDiffusion, which leverage the pretrained CLIP model to provide multimodal text-image representations and strong image generation capabilities. On the FFHQ dataset, our model outperformed previous state-of-the-art methods by 4.4\% in clipscore and generated very realistic images even when the text was both in and out of distribution. The pretrained models and codes will soon be available at \url{https://github.com/INFINIQ-AI1/CLIPVQDiffusion}

\keywords{Langauge Free training \and CLIP \and VQ-Diffusion}
\end{abstract}

\begin{figure}[!t]
    \centering
    \includegraphics[width=0.9\columnwidth]{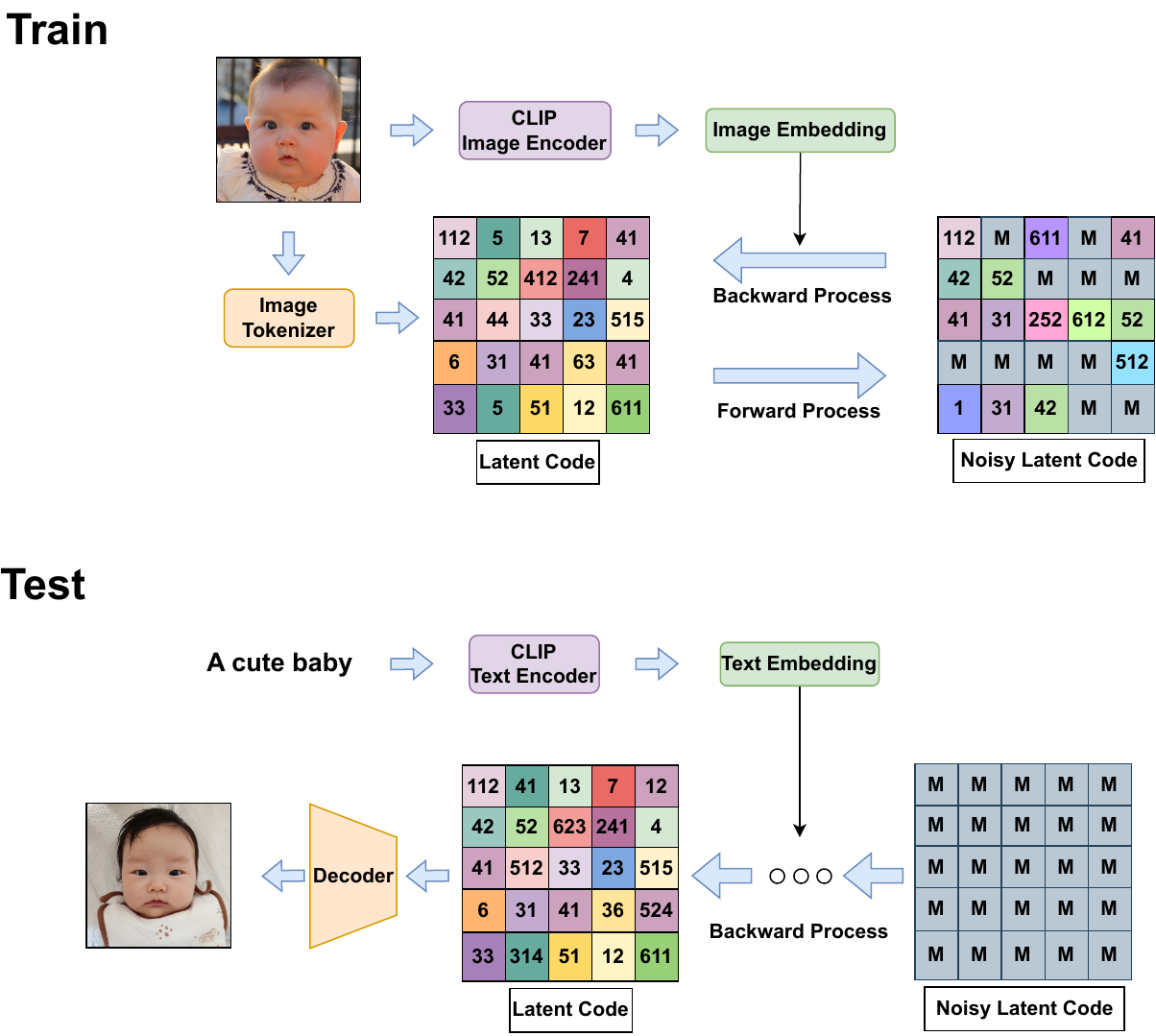}
    \caption{An overview of our CLIP-VQDiffusion approach. At the training stage, we embed input image to clip image embedding space, and also get clean latent code using image tokenizer. Conditioned on image embedding, our vector quantized diffusion model restore noisy latent code to clean latent code. At the inference stage, instead of the image embedding, CLIP text embedding is used as condition of our diffusion model to generate corresponding latent code.}
    \label{fig:clipvqdiffusion}
\end{figure}

\section{Introduction}

Conditional generative method model the joint probability distribution of input variable with output data given condition. This condition could include sketches, segmentations, image edge features, or various other conditions, including text and sounds\cite{ldm,soundtoimage} . Especially, text conditional image generation aims to generate image well-aligned with given texts. Since text is easier to use as condition and could contain rich information, it has attracted the interest of many researchers. 

Earlier studies of text conditional image generation\cite{dmgan, dfgan, attngan} directly generated pixels from given text embeddings. However, generating images using pixel space could incur high computational costs and often worked within limited domains. To generate perceptually meaningful images in general domains with low computational costs, recent studies have typically generate latent vectors instead of raw pixels from text embeddings\cite{cameleon, glide, magvitv2, ldm, vqd, parti} Recently, languauge models that receive text tokens and generate visual tokens have shown promising results in this field\cite{parti, cameleon, magvitv2}. Alongside this line of research, diffusion models also generate latent vectors or codes using text embeddings\cite{ldm, vqd}. However, to train a text conditional image generation model, a large amount of text paired image datasets is still required. 

Large scale datasets \cite{laion400m, laion5b, jft4b, align} boosted text conditional image generation quality. However, in some domains it could be difficult to make such datasets and usually it could be costly. Also, famous face datasets\cite{ffhq,celeba,vggface,millionceleb} don't have descriptions corresponding to the face but only attribute annotations. In this situation, training text conditional image generation model could be difficult. Leveraging CLIP\cite{clip} could be a solution in this situation. CLIP is trained on millions of general text-image pairs, so it could give vision-language multimodal representations which relate two modalities. Leveraging CLIP model, we could train text to image generation model with only image datasets without text pairs. Figure 1 shows our CLIP-VQDiffusion network. 

We leverage multimodal embedding space of CLIP to use relation of vision-language space. In the training stage, we use image embedding as condition of generative models. Since CLIP image embedding and corresponding text embedding is located near in the embedding space, we could use text embedding in the inference stage to get expected image. In the inference stage, the code starts from the fully masked code and reach to clean latent code. 

Our contributions of this paper are 

\begin{itemize}
    \item We propose a text conditional image generation model which leverage CLIP and vector quantized diffusion model to train using image datasets without texts.
    \item Throungh both qualative and quantitive evaluation on FFHQ and COCO datasets, we demonstrate our model could generate text aligned image in languauge free training method. 
\end{itemize}

\section{Related Works}

\subsection{Diffusion Models}
With large language models, Diffusion models gained great success in text to image generation tasks. Diffusion model resembles Hierarchical Variational Auto Encoder(HVAE). Compared with HVAE, Diffusion model assumes forward process as diffusion process, which means some Gaussian noise is added to input, and result in pure Gaussian noise after total timesteps. Diffusion model is known to converge stably, and could generate diverse and photo-realistic image. Recently, Diffusion models also incorporate visual tokenizer as the input space to reduce computational costs and increase visual perceptuality. Rombach et al\cite{ldm} proposed operating diffusion process in the continuous latent space. Gu et al\cite{vqd} proposed vector quantized diffusion model which use discrete latent space for diffusion process, and gained good results in the image generation tasks. In this work, we used vector quantized diffusion model\cite{vqd} and gained perceptually meaningful image given text.

\subsection{Language Free Training}
There are few language free training methods. As far as I know, every methods leveraged CLIP multimodal embedding power to train using only images. Lafite\cite{lafite} leveraged CLIP and StyleGAN\cite{stylegan} decoder to generate image. clip2latent\cite{clip2latent} leveraged pretrained StyleGAN decoder and CLIP model. By using pretrained StyleGAN decoder, and CLIP model, clip2latent make it possible to train text conditional image generatoin model without images and texts. ClipGen\cite{clipgen} adopted transformer decoder to generate image. Since autoregressive transformers generate image token from left to right, it have accumulation of errors problems. However, our structure is free from accumulation of error problems since our model use diffusion model as image generator. Also, we adopted more geneally used vector qunatized variational autoencoder instead of StyleGAN decoder to decoder image from predicted code.

\section{Background}
\subsection{Vector Quantized VAE}
Recently, Many generative models use visual tokenizer. Variational Autoencoder(VAE) can be used as visual tokenizer and it could reduce dimension of pixel space. We used Vector Qunatized Variational Autoencoder(VQ-GAN)\cite{vqgan} to learn photo-realistic and perceptually rich codebook $\mathcal{Z} = \{z\}^{K}$. After pretraining VQ-GAN, We use encoder $E$ and codebook $\mathcal{Z}$ to encode image into discrete latent code. and decoder $G$ is used to generate image given predicted discrete latent code. \ 

We used Gumbel softmax training method\cite{gumbel_softmax} for training our VQ-GAN. This method uses Gumbel softmax which is defined as 
\begin{equation}
    y_{i} = {{\exp((\log(\pi_{i})+g_{i})/\tau)}\over{\sum_{j=1}^{k} \exp((\log(\pi_{j})+g_{j})/\tau)}} \quad \textnormal{ for } i = 1,...,K
\end{equation}
where $\tau$ and $\pi$ stand for temperature and parameters respectivly, and $g_{1}, ... g_{K}$ are sampled from Gumbel(0,1) distribution. The Gumbel-softmax distribution is smooth for $\tau > 0$ making it possible to calculate gradient with respect to parameter $\pi$. In the training VQ-GAN, using this distribution, we get image latent as $\sum_{i=1}^{K} y_{i}z_{i}$ for each resolution. And the temperature $\tau$ anneals from high temperature to low temperature as training proceed. Therefore, logit value changes from less variance to large variance. When used in the inference stage, we select most high probability index and quantize image into discrete token. This stochastic quantization method is used to avoid codebook collapse and mitigate inference misalignment\cite{reg_vqvae}.

\subsection{Vector Quantized Diffusion models}
When raw image $x \in \mathbb{R}^{3 \times H \times W}$ is given, pretrained discrete visual tokenizer create visual token $x_{0} \in \mathbb{Z}^{H/f \times W/f}$ where $f$ refers the compression rate. Conditional diffusion model is designed to model conditional distribution $q(x|y)$. Forward process in diffusion model corrupt latent gradually using fixed Markov chain $q(x_{t}|x_{t-1})$. When latent data $x_{t}$ is continuous, forward process is usually defined as $x_{t} = \sqrt{1-\beta_{t}}x_{t-1}+\sqrt{\beta_{t}}\mathcal{N}(0,I)$. After total timesteps $T$, latent data become pure noise $X_{T}$. If $x_{0}$ is known, one could calculate posterior distribution $p(x_{t-1}|x_{t},x_{0})$. In reverse process, model predict $x_{0}$ or added noise given $x_{t}, t$ and condition $c$. However, when latent space is discrete space, one cannot corrupt image by adding Gaussian noise in the latent vector. vector quantized diffusion model\cite{vqd}  corrupt discrete latent data $x_{0}$ using mask and replace strategy using transition matrix $Q_{t}$ where $[Q_{t}]_{mn} \coloneq q(x_{t}=m|x_{t-1}=n)$ and formulated as below \\

\begin{equation}
Q_{t} = \begin{bmatrix}
\alpha_{t}+\beta_{t} & \beta_{t} & \beta_{t} & \cdots & 0 \\
\beta_{t} & \alpha_{t}+\beta_{t} & \beta_{t} & \cdots & 0 \\
\beta_{t} & \beta_{t} & \alpha_{t}+\beta_{t} & \cdots & 0 \\
\vdots & \vdots & \vdots & \ddots & & \\
\beta_{t} & \beta_{t} & \beta_{t} & \cdots & 0 \\ 
\gamma_{t} & \gamma_{t} & \gamma_{t} & \cdots & 1  
\end{bmatrix}
\end{equation}

Since the property of Discrete time Markov chain, one can easily derive few step forward equation 
\begin{equation}
q(x_{t}|x_{0})=v(x_{t})^{T}Q_{t}Q_{t-1}\cdots Q_{1}v(x_{0})=v(x_{t})^{T}\overline{Q}_{t}v(x_{0})
\end{equation}
where $\overline{Q}_{t}=Q_{t}Q_{t-1}\cdots Q_{1}$ and $v(x_{t})$ is one-hot column vector with column length $K$. Like continuous diffusion case, given $x_{0}$, one can calculate reverse process $q(x_{t-1}|x_{t},x_{0})$ in tractable form. 
\begin{equation}
q(x_{t-1}|x_{t},x_{0}) = {{q(x_{t-1}|x_{0})q(x_{t}|x_{t-1})}\over{q(x_{t}|x_{0})}} = {{{v(x_{t-1})^{T}\overline{Q}_{t-1}v(x_{0})\cdot v(x_{t})^{T}Q_{t}v(x_{t-1})}}\over{v(x_{t})^{T}\overline{Q}_{t}v(x_{0})}}
\end{equation}
To training reverse diffusion process, vector quantized diffusion model learn conditional denoising process $p_{\theta}(x_{t-1}|x_{t},y)$ to predict $q(x_{t-1}|x_{t},x_{0})$. the training objective of vector qunatized diffusion model is variational lower bound which is stated as below. \

\begin{align}
\mathcal{L}_{vlb} &= \mathcal{L}_{0} + \mathcal{L}_{1} + \cdots + \mathcal{L}_{T} \\
\mathcal{L}_{0} &= -\log{p_{\theta}(x_{0}|x_{1},y)} \\
\mathcal{L}_{t-1} &= D_{KL}(q(x_{t-1}|x_{t},x_{0})\| p_{\theta}(x_{t-1}|x_{t},y)) \\
\mathcal{L}_{T} &= D_{KL}(q(x_{T}|x_{0})\|p(x_{T}))
\end{align}

\section{CLIP-VQDiffusion} 
\begin{figure}[h]
    \centering
    \includegraphics[width=1\columnwidth]{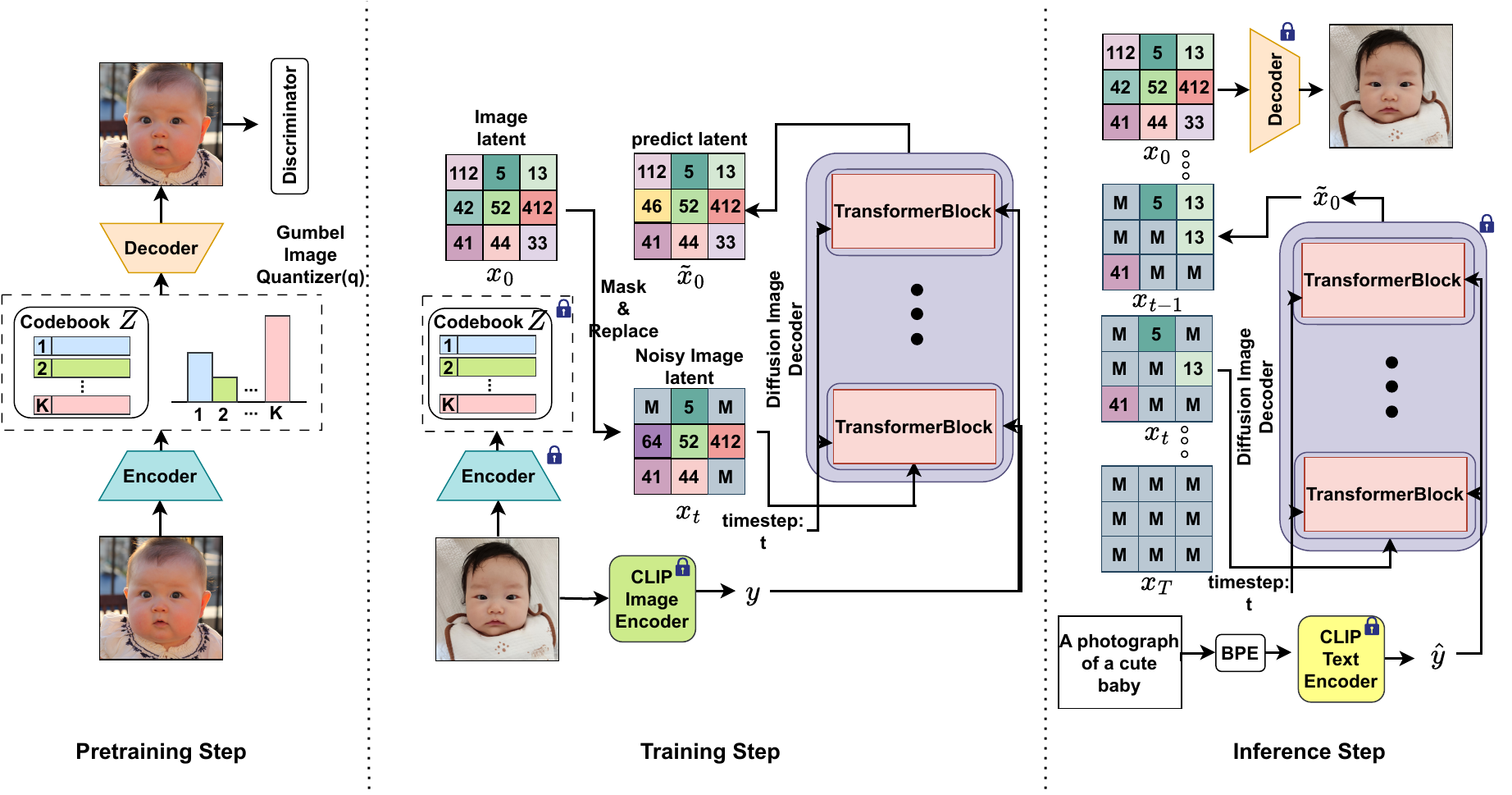}
    \caption{Structure of our model. In the pretraining step, we train visual tokenizer using Gumbel softmax training method. In the training Step, we encode image using visual tokenizer to get image latent code. Then Diffusion Image Decoder learn to predict image latent given noisy image latent and timestep, and CLIP image embedding. At the inference step, we generate image latent from all masked latent code, using CLIP text embedding.}
    \label{fig:clipvqdiffusion_whole}
\end{figure}

Figure \ref{fig:clipvqdiffusion_whole} shows whole structure of our proposed model. In the pretraining stage, we train our VQ-GAN model by reconstructing photo-realsitic image. In this stage, we train image quantizer using Gumbel softmax training method\cite{gumbel_softmax}. In the training step, we tokenize given image $x$ to latent code $x_{0}$. Then, we could get noisy latent $x_{t}$ by $q(x_{t}|x_{0})$ and train our model to denoise noisy latent $x_{t}$ to clean latent code $x_{0}$ conditioned on CLIP image embedding. 

We could use Gaussian noise to CLIP image embedding to connect properly between image embedding and text embedding as proposed in Lafite\cite{lafite}. After training, we finetune the model to predict more correct image by sacrificing diversity using classifier free guidance method\cite{improved_vqd}. After finetuning step, we could generate image using text prompt. 

\begin{wrapfigure}{r}{0.4\textwidth}
\vspace{-10pt}
\begin{center}
\includegraphics[width=0.8\linewidth]{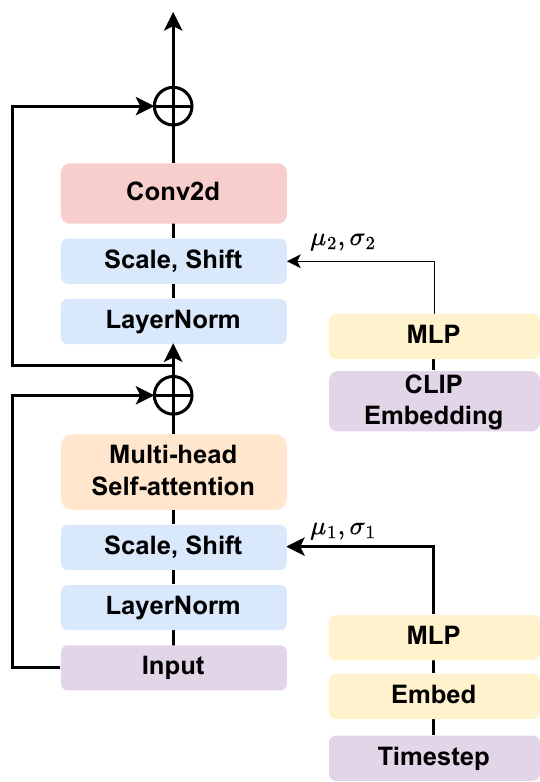} 
\vspace{0pt}
\caption{Structure of transformer block incorporating CLIP embedding}
\label{fig:block}
\end{center}
\vspace{-60pt}
\end{wrapfigure}

Since CLIP model could connect text embedding $\hat{y}$ and its corrsponding image embedding $y$ in similar multimodal embedding space, we could generate image using text embedidng.

\subsection{Adalayer-CLIP}
Figure \ref{fig:block} shows transformer block in the diffusion model which incorporate CLIP embedding. It embeds CLIP embedding $e$ to get scale and shift $\sigma_{e}, \mu_{e}$ through multi-layer perceptrons. Then we use scale and shift to change the distribution of output according to CLIP embedding by Adaptive Layer Norm method given by 
\begin{equation}
    \text{AdaLN}(x) = \sigma_{e} \text{LN}(x) + \mu_{e} 
\end{equation}

\subsection{Pseudo Text Embedding} 

CLIP model is known to connect image and corresponding text caption in multimodal space. Some researchers pointed that there are gap between image embedding and corresponding text embedding\cite{mindthegap}. Actually, image embedding and its corresponding text embedding usually have low cosine similarity(around 0.3 $\sim$ 0.4). 

\begin{wrapfigure}{r}{0.4\textwidth}
\vspace{-20pt}
\begin{center}
\includegraphics[width=0.7\linewidth]{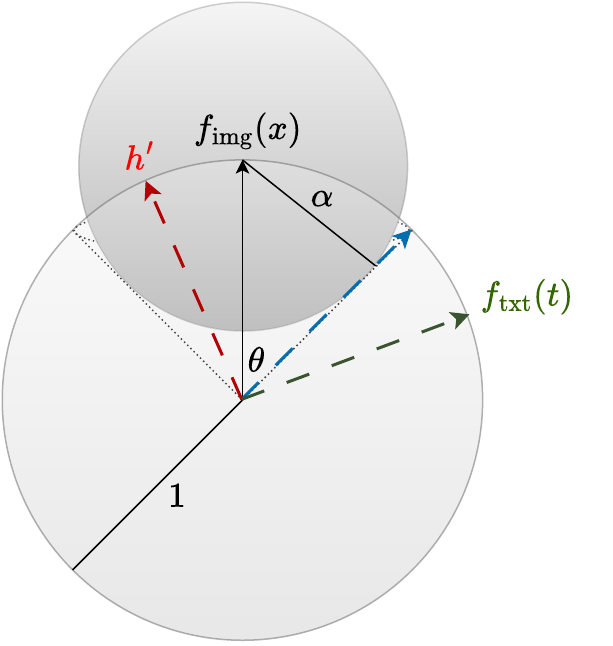} 
\vspace{10pt}
\caption{Explanation of Pseudo text embedding. Since CLIP image embedding $f_{\text{img}}(x)$ and text embedding $f_{\text{txt}}(t)$ located far from each other, we could add Gaussian noise to $f_{\text{img}}(x)$ and get pseudo text embedding $h'$ in the training step}
\vspace{-20pt}
\label{fig:pseudotextembedding}
\end{center}
\vspace{-50pt}
\end{wrapfigure}

When we use normalized CLIP embedding, it means the angle between two embedding is over 60$^{\circ}$ which could be too far to connect two embeddings. While some researchers used CLIP image embedding directly to the conditional image generation model\cite{clipgen}, other researchers added Gaussian noise to image embedding $f_{\text{img}}(x)$ and normalized it to overcome this gap\cite{lafite, clip2latent}, and calculated pseudo text embedding $h'$ by 
\begin{align}
h &= f_{\text{img}}(x) + \alpha * ( {{e}\over{\|e\|}}) \\
h'&=  {{h}\over{\|h\|}}.
\end{align}
Figure \ref{fig:pseudotextembedding} shows pseudo text $h'$. According to given hyper parameter $\alpha$, the maximum degree $\theta$ can be calculated as $\theta = \arcsin(\alpha)$. 

\subsection{Classifier-free guidance}
After learning denoising process, we finetune the diffusion image decoder using classifier free guidance\cite{improved_vqd}. For classifier free guidance, we change the CLIP image embedding with learnable parameters of same dimension with probability $p$. and in the inference step, we calculated the target logit with equation below. 
\begin{equation*}
    \log p_{\theta}(x_{t-1}|x_{t},\hat{y}) = \log p_{\theta}(x_{t-1}|x_{t},\hat{y})+(s-1)(\log p_{\theta}(x_{t-1}|x_{t},\hat{y}) - \log p_{\theta}(x_{t-1}|x_{t}, y')) 
\end{equation*}
where $\hat{y}$ and $y'$ stands for the CLIP text embedding and learned parameters respectively, and $s$ stands for the guidance scale hyper parameter. 

\section{Experiments}
In this section, we explain how we evaluate our models comparing with other methods. first we explain datasets we used to evaluate our models and evaluation metrics. next we compare our model with other text to image generation models. Here, we compare our model with other model which is trained under language free training setup.

\subsection{Datasets}
We trained and evaluated our model using datasets commonly used in image generation tasks which are COCO, FFHQ datasets.
\subsubsection{MS-COCO} We used 2014 split of coco datasets\cite{mscoco}. it contains 80k images for training and 40k test set images. Each image include 5 text captions describing the image. We used complete image sets to train our visual tokenizer and diffusion image decoder. We used only text from validation datasets for quantitive evaluations and visual results. 
\subsubsection{FFHQ} FFHQ dataset\cite{ffhq} contains 70k image datasets with its attribute labeled without train validation split. We trained our visual tokenizer and image generator using whole datasets.\\

\subsection{Evaluation metrics}
To evaluate the quality of our generated image quantitively, we used commonly used metrics to evaluate image. We used Frechet Inception Distance(FID)\cite{fid} to evaluate our image. FID computes the Frechet Distance between the distribution of features extracted from inception model. To calculate FID score, we randomly selected 30000 text captions in validation datasets. Then we generated images using these text captions and calculated frechet distance between synthetic images and real world images. For statistics of real world images, we used whole validation images. Some researchers used center cropped images while others squeezed validation images. Here, we used squeeze method to calculate real world statistics. we also used clipscore which calculate cosine similarity between input texts and generated images, and Inception Score(IS) to measure quality and diversity of generated images. We used validation caption for text prompts for evaluating COCO datasets. For FFHQ datasets, we used 64 text prompts used in clip2latent\cite{clip2latent}. As in clip2latent, we generated 16 candidates for each text prompts and selected best image using cosine similarity, and calcuated clip score in the selected images. 

\subsection{Implementation details}
\subsubsection{Pretrained Visual Tokenizer} 
In the pretraining stage, we trained VQ-GAN on COCO and FFHQ datasets. All models are trained in the same hyper-paramters and architecture. We set codebook number $|\mathcal{Z}|=4096$, embedding size $dim_{z}=256$, and sequence length $|S|=16\times 16=256$, and compression rate $f=16$. when training, 3-layer PatchGAN\cite{patchgan} is used as a Discriminator while Discriminator is not used until Generator trained 25000 steps. Adam optimimzer is used with $(\beta_{1}, \beta_{2})=(0.5,0.9)$, and learning rate $0.0000045$ adjusting linearly by batch size. For Gumbel quantizer, we anneled temperature from 0.9 to 0.01. we trained each datasets for 100 epochs with batch size 32. 

\subsubsection{Diffusion Image Decoder} 
We trained vector quantized diffusion model with same hyper parameters and architectures for COCO and FFHQ datasets. The transformer architecture of Diffusion model's inner dimension is 512. and it have 16 heads consist of 24 layers. For the other hyper-parameters, See Section 5.5 for more details.

\subsection{Compare with state of the arts}

\begin{figure}[t]
    \centering
    \includegraphics[width=1\columnwidth]{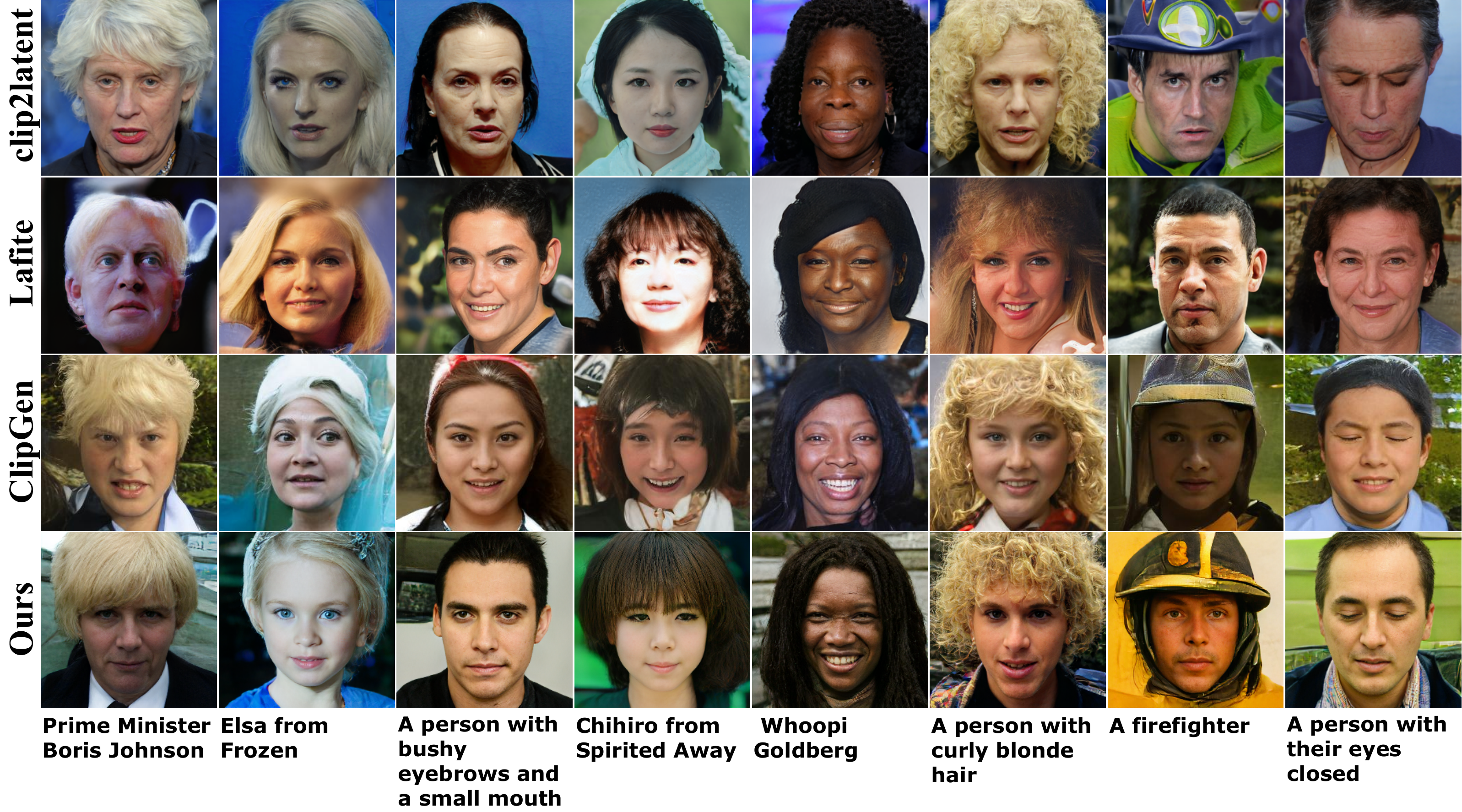}
    \caption{Sample images generated from our method, clip2latent, Lafite and Clipgen. our method achieves high quality sample with great details. all images generated with fixed text prompt "A photograph of".}
    \label{fig:our_ffhq}
\end{figure}

\begin{wrapfigure}{t}{0.5\textwidth}
\vspace{-30pt}
\includegraphics[width=1\linewidth]{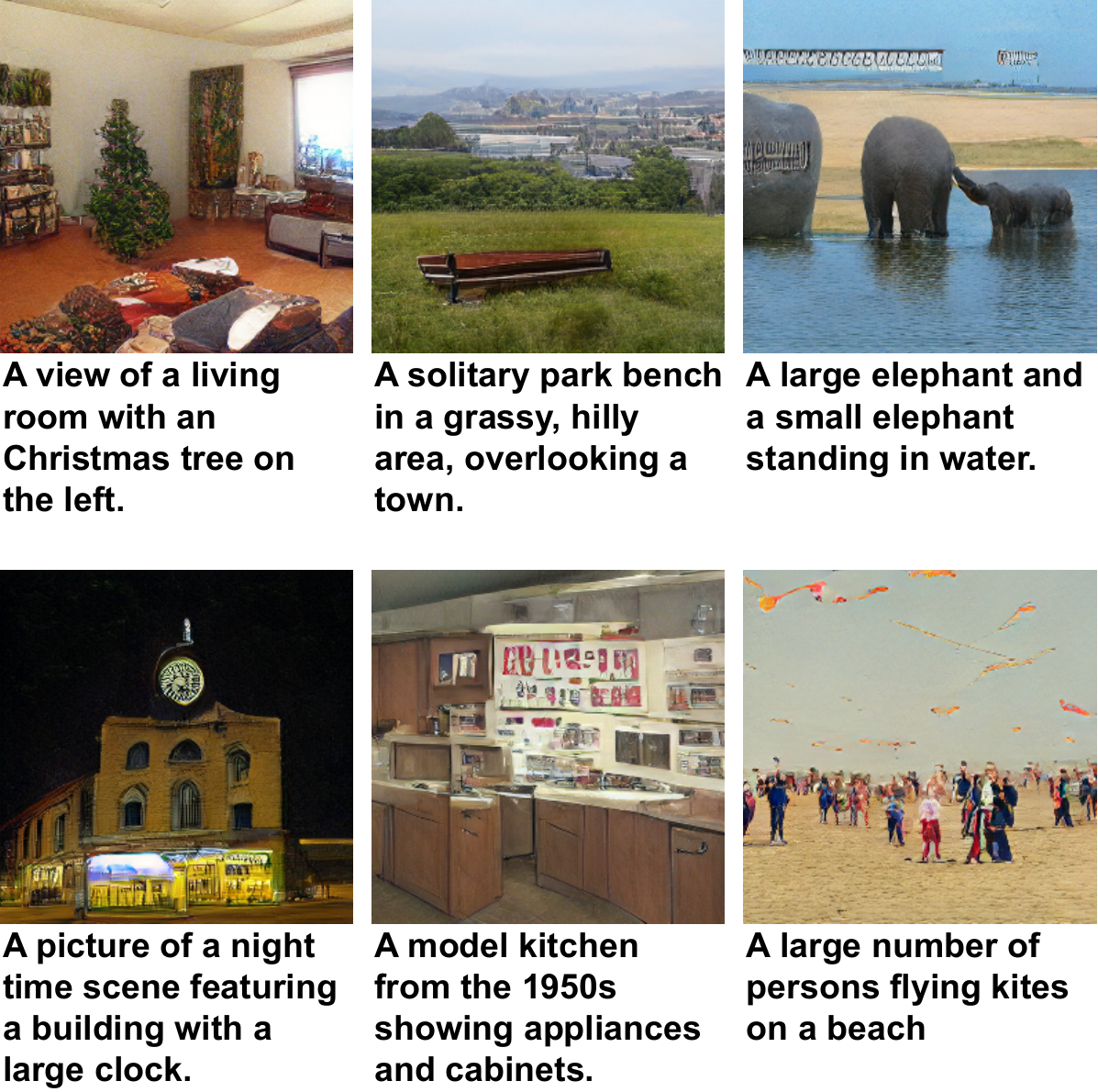} 
\vspace{-20pt}
\caption{Sample Images generated from our model trained on COCO datasets}
\vspace{-30pt}
\label{fig:our_coco}
\end{wrapfigure}

Our model create high quality image with only training images without paired texts. In this section we present generated images using our models. Figure \ref{fig:our_ffhq} compares our model trained on FFHQ datasets with clip2latent\cite{clip2latent}, LAFITE\cite{lafite}, ClipGen\cite{clipgen}.

 We use the text-free MM-CelebA trained model by the origianl authors of LAFITE when evaluating Lafite. For all models, we selected 16 candidates for each text prompts and selected best matching image using clipscore. Our model generates greatly detailed images well aligned with text prompts. Especially, our model excels in generating image with texts which is out of distribution(Prime Minister Boris Johnson, Elsa from Frozen, Whoopi Goldberg, etc) and achieved best clipscore, See table \ref{tab:ffhq_coco_result}.

 We also show our result on MS-COCO datasets. See Figure \ref{fig:our_coco}. our model generate images corresponding to text prompts and achieved comparable result with other text conditional image generation model without texts. See table \ref{tab:ffhq_coco_result} for the result. 

\begin{table}[h]
\vspace{-20pt}
\caption{Result of FFHQ and COCO Image generation}
    \centering
    \begin{tabular}{C{2cm}|L{2.5cm}|C{1.5cm}C{1.5cm}C{1.5cm}}
        \hline
        Dataset & Model & CLIP $\uparrow$ & FID $\downarrow$& IS $\uparrow$\\ 
        \hline
        FFHQ &  clip2latent\cite{clip2latent} & 0.316 & - & - \\
        & ClipGen\cite{clipgen} & 0.29 & - & - \\ 
        & LAFITE\cite{lafite} & 0.278 & - & - \\ 
        & Ours & 0.330 & - & - \\ 
        \hline 
        COCO & ClipGen\cite{clipgen} & 0.283 & 22.99 & 17.53 \\ 
        & LAFITE\cite{lafite} & 0.322 & 18.69 & 26.97\\ 
        & Ours & 0.295 & 27.63 & 19.27 \\
        \bottomrule
    \end{tabular}
    \vspace{-20pt}
    \label{tab:ffhq_coco_result}
\end{table}

\subsection{ablation studies}

\subsubsection{Hyper parameter search} We analyzed the effect of 3 hyper parameters here, which are guidance scale $s$ for classifier free guidance, Gaussian noise scale $\alpha$ used to CLIP embedding and truncation ratio $r$. Table \ref{tab:coco_hyperparam} shows the ablation studies on truncation ratio and guidance scale. Guidance scale $s=1.0$ means there are classifier free guidance, so not fine-tuned model. This table is for the case where Gaussian noise scale $\alpha = 0.25$ since it performed best in COCO dataset. we chose truncation rate $r=0.9$ and guidance scale $s=3$ for COCO dataset. As one can see, there are trade-off between FID and IS score. In FFHQ dataset, we selected $r=0.85$ and $s=1.15$. 

\begin{table}[h]
\vspace{-20pt}
\caption{ablation studies on guidance scale and truncation ratio on COCO datasets, the score is clipscore, FID, IS}
\centering
\begin{tabular}{|C{0.5cm}|C{0.5cm}|C{2.6cm}|C{2.6cm}|C{2.6cm}|C{2.6cm}|}
\hline
\multirow{9}{*}{\rotatebox{90}{\!\!\!\!\!\!\!\!\!\!\!\!\!\! Guidance scale}}&
\multicolumn{5}{c|}{Truncation ratio} \\
\cline{1-6}
&  & 0.75 & 0.8 & 0.85 & 0.9\\
\cline{2-6}
&1.0  & 0.272\slash33.54\slash15.79 & 0.273\slash32.58\slash15.80 & 0.273\slash32.17\slash15.47 & 0.272\slash32.48\slash15.26 \\
\cline{2-6}
  & 1.1 & 0.277\slash31.37\slash16.74 & 0.278\slash30.26\slash16.80 & 0.278\slash29.69\slash16.73 & 0.277\slash29.96\slash16.43 \\
\cline{2-6}
  & 1.2 & 0.280\slash30.72\slash17.23 & 0.281\slash26.69\slash17.27 & 0.280\slash28.76\slash17.32 & 0.280\slash29.05\slash16.82 \\
\cline{2-6}
  & 1.3 & 0.282\slash30.42\slash17.47 & 0.282\slash29.38\slash17.54 & 0.283\slash28.36\slash17.52 & 0.282\slash28.54\slash17.34 \\
\cline{2-6}
  & 1.5 & 0.285\slash29.76\slash18.07 & 0.286\slash29.04\slash18.08 & 0.286\slash27.97\slash18.22 & 0.286\slash27.83\slash17.81 \\
\cline{2-6}
  & 2 & 0.290\slash29.70\slash18.49 & 0.291\slash28.61\slash18.79 & 0.291\slash27.83\slash18.87 & 0.291\slash\underline{27.24}\slash18.83 \\
\cline{2-6}
  & 3 & 0.294\slash29.71\slash19.01 & 0.294\slash28.95\slash19.09 & \underline{0.295}\slash28.41\slash\underline{19.43} & \underline{0.295}\slash27.63\slash19.27 \\
\cline{2-6}
  & 5 & 0.289\slash29.88\slash18.16 & 0.289\slash30.21\slash17.90 & 0.288\slash30.83\slash18.02 & 0.287\slash31.58\slash17.67 \\
\hline
\end{tabular}
\label{tab:coco_hyperparam}
\end{table}

\begin{wrapfigure}{t}{0.5\textwidth}
\vspace{-20pt}
\includegraphics[width=1\linewidth]{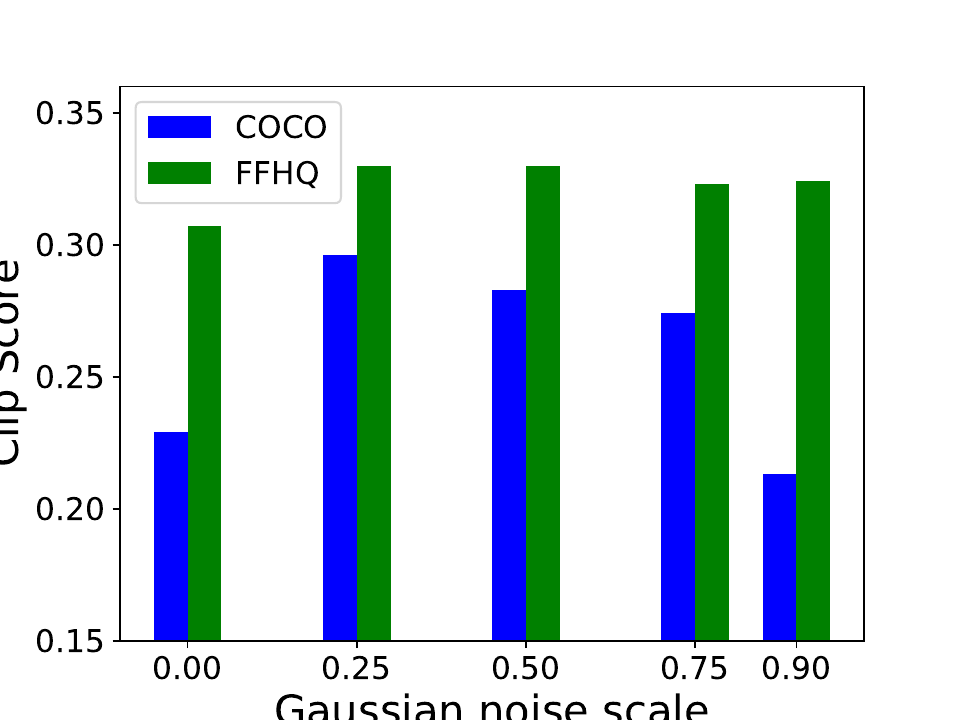} 
\vspace{-10pt}
\caption{CLIP score of COCO, FFHQ datasets under different Gaussian noise scales}
\vspace{-15pt}
\label{fig:test_alpha}
\vspace{-30pt}
\end{wrapfigure}

 We also tested on the effect of Gaussian noise scale $\alpha$ in Figure \ref{fig:test_alpha}. For all COCO and FFHQ datasets, Gaussian noise added to CLIP image embedding helped connecting image embedding and text embedding in inference step. noise scale $\alpha = 0.25$ achieved the best CLIP score in both datasets and other metrics for COCO datasets.

 In the case of COCO, $\alpha=0.25$ achieved better result with large margin compared with other $\alpha$. However, in the case of FFHQ, it is slightly better and low $\alpha$ cause image to have artifacts since it reduces generalization power. So we chose $\alpha=0.5$ for the FFHQ case. 

\subsection{Prompts}
We use validation caption annotation for COCO datasets. However, Since FFHQ dataset does not have caption on the image, we used 64 prompts proposed in clip2latent\cite{clip2latent}. We sampled 16 images for each prompts and selected one image per prompts which have maximum clipscore in the samples. We list the 64 prompts used to evaluate model trained on FFHQ datasets. 

\label{app:prompts}
\tiny
\begin{multicols}{3}
\begin{enumerate}
\item a person with glasses
\item a person with brown hair
\item a person with curly blonde hair
\item a person with a hat
\item a person with bushy eyebrows and a small mouth
\item a person smiling
\item a person who is angry
\item a person looking up at the sky
\item a person with their eyes closed
\item a person talking
\item a man with a beard
\item a happy man with a moustache
\item a young man
\item an old man
\item a middle aged man
\item a youthful man with a bored expression
\item a woman with a hat
\item a happy woman with glasses
\item a young woman
\item an old woman
\item a middle aged woman
\item a baby crying in a red bouncer
\item a child with blue eyes and straight brown hair in the sunshine
\item an old woman with large sunglasses and ear rings
\item a young man with a bald head who is wearing necklace in the city at night
\item a youthful woman with a bored expression
\item President Xi Jinping
\item Prime Minister Boris Johnson
\item President Joe Biden
\item President Barack Obama
\item Chancellor Angela Merkel
\item President Emmanuel Macron
\item Prime Minister Shinzo Abe
\item Robert De Niro
\item Danny Devito
\item Denzel Washington
\item Meryl Streep
\item Cate Blanchett
\item Morgan Freeman
\item Whoopi Goldberg
\item Usain Bolt
\item Muhammad Ali
\item Serena Williams
\item Roger Federer
\item Martina Navratilova
\item Jessica Ennis-Hill
\item Cathy Freeman
\item Christiano Ronaldo
\item Elsa from Frozen
\item Eric Cartman from South Park
\item Chihiro from Spirited Away
\item Bart from the Simpsons
\item Woody from Toy Story
\item a university graduate
\item a firefighter
\item a police officer
\item a butcher
\item a scientist
\item a gardener
\item a hairdresser
\item a man visiting the beach
\item a woman giving a TED talk
\item a child playing with friends
\item a person watching birds in the forest
\end{enumerate}
\end{multicols}
\normalsize

\section{Conclusion}

In this paper, we propose novel langague free training model leveraging CLIP and vector quantized diffusion model. Our model achieved best score in FFHQ datasets under langauge free training setting, and comparable result in the COCO dataset also.

\begin{figure*}[t!]
        \caption{Generated samples from our model trained on FFHQ datasets}
	\centering
	\includegraphics[width=1.0\linewidth]{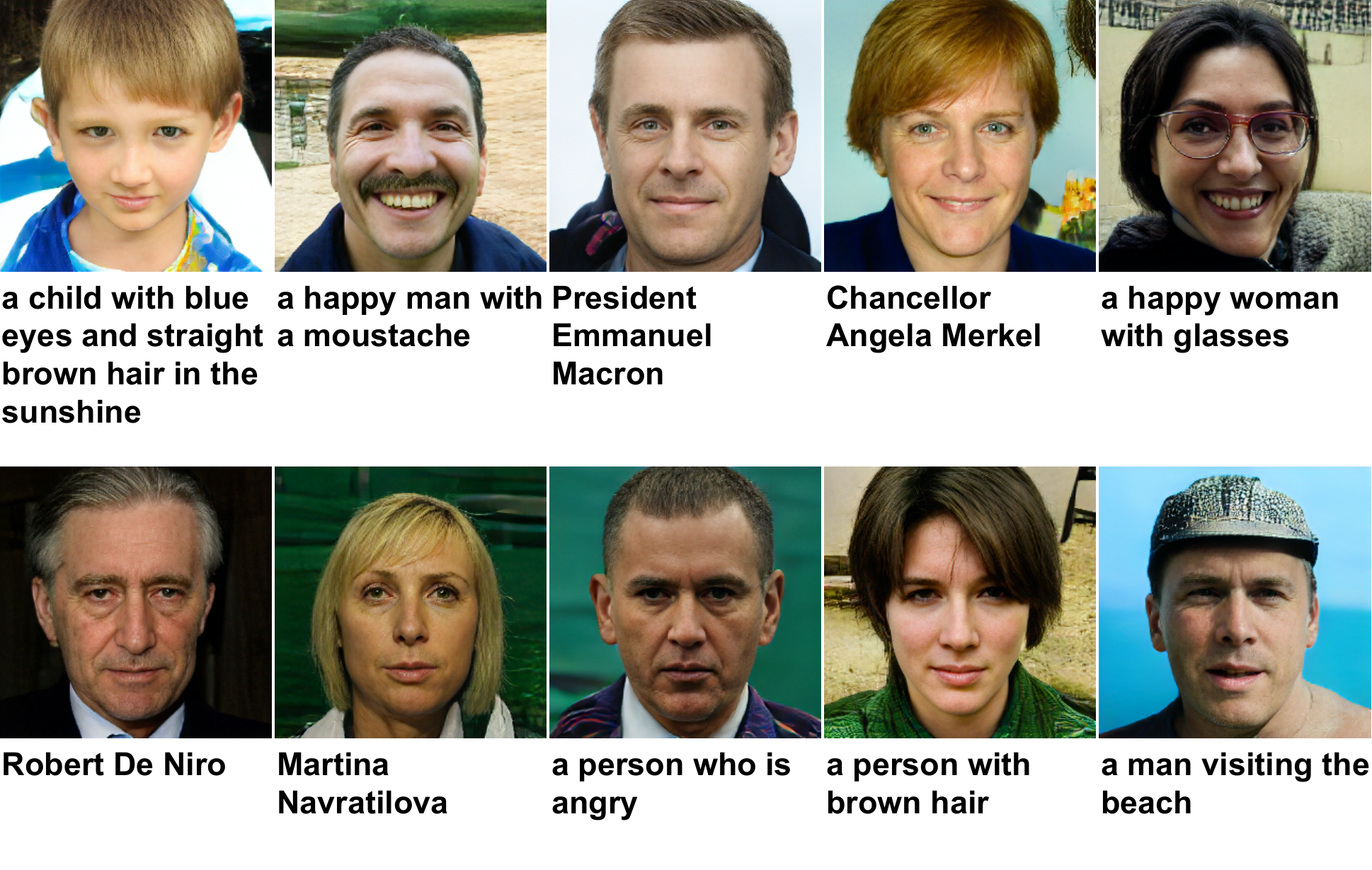}
	
	\label{fig:suppcom}
\end{figure*}

\begin{figure*}[h]
        \caption{Generated samples from our model trained on COCO datasets}	
        \centering
	\includegraphics[width=1.0\linewidth]{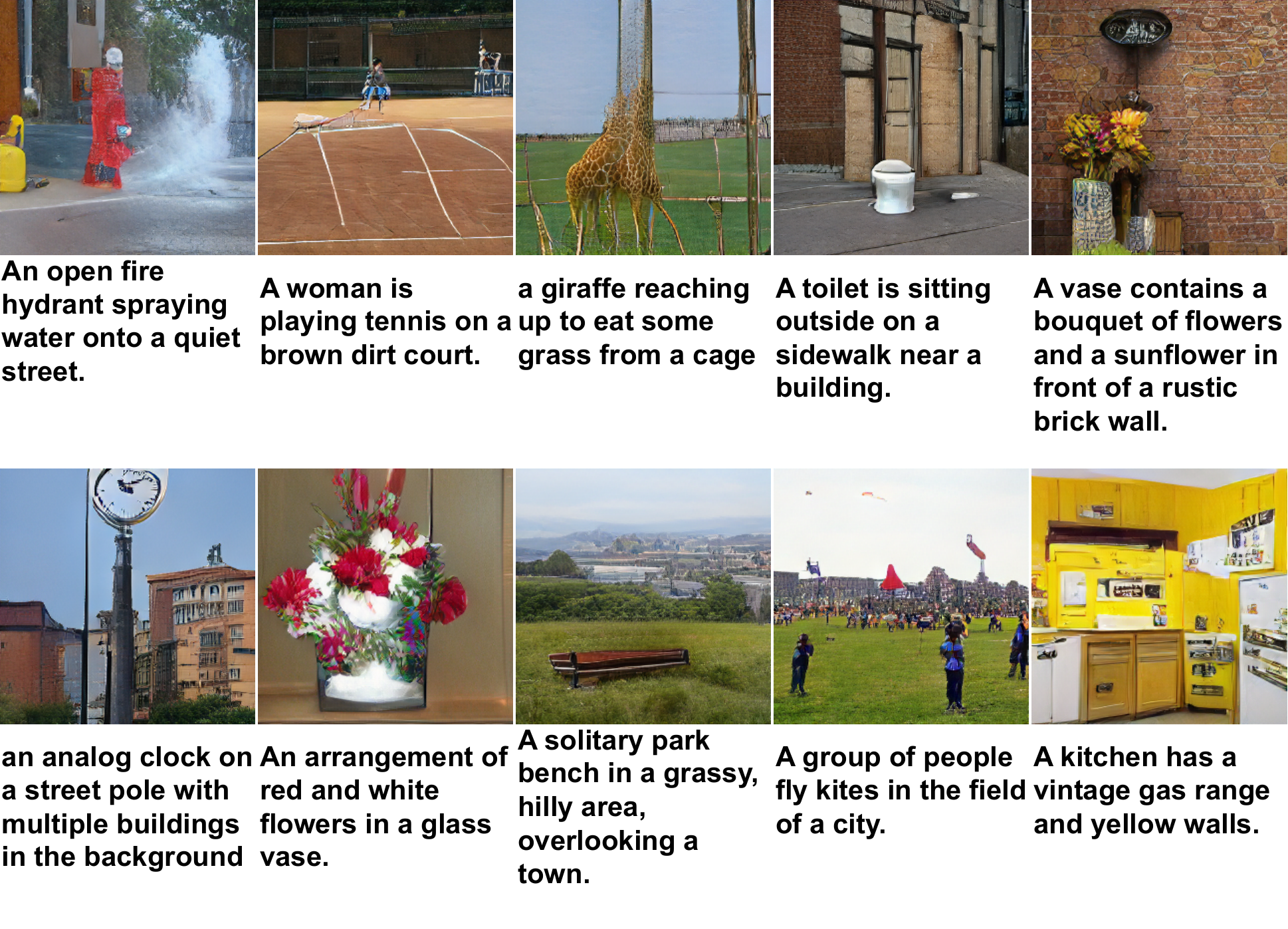}
	\label{fig:ffhq}
\end{figure*}

\bibliographystyle{splncs04}
\bibliography{references}
\end{document}